\documentclass[letterpaper, 10 pt, conference]{ieeeconf}  

\IEEEoverridecommandlockouts                              

\usepackage{graphics} 
\usepackage{epsfig} 
\usepackage{mathptmx} 
\usepackage{times} 
\usepackage{amsmath} 
\usepackage{amssymb}  

\title{\LARGE \bf
MIXPINN: Mixed-Material Simulations by Physics-Informed Neural Network                                     
}

\author{Xintian Yuan$^{1,2}$, Yunke Ao$^{1,2,3}$, Boqi Chen$^{1,3}$, Philipp Fürnstahl$^{2,3}$
\thanks{$^{1}$ Department of Computer Science,
        ETH Zurich, Universitätstrasse 6, 8092 Zürich, Switzerland
        {\tt\small xinyuan@ethz.ch}}%
\thanks{$^{2}$ ROCS, Balgrist University Hospital, University of Zurich, Forchstrasse 340, 8008 Zürich, Switzerland}%
\thanks{$^{3}$ ETH AI Center, ETH Zürich, Andreasstrasse 5, 8092 Zürich, Switzerland}%
        }

\begin{document}

\maketitle
\thispagestyle{empty}
\pagestyle{empty}

\begin{abstract}
Simulating the complex interactions between soft tissues and rigid anatomy is critical for applications in surgical training, planning, and robotic-assisted interventions. Traditional Finite Element Method (FEM)-based simulations, while accurate, are computationally expensive and impractical for real-time scenarios. Learning-based approaches have shown promise in accelerating predictions but have fallen short in modeling soft-rigid interactions effectively. We introduce MIXPINN, a physics-informed Graph Neural Network (GNN) framework for mixed-material simulations, explicitly capturing soft-rigid interactions using graph-based augmentations. Our approach integrates Virtual Nodes (VNs) and Virtual Edges (VEs) to enhance rigid body constraint satisfaction while preserving computational efficiency. By leveraging a graph-based representation of biomechanical structures, MIXPINN learns high-fidelity deformations from FEM-generated data and achieves real-time inference with sub-millimeter accuracy. We validate our method in a realistic clinical scenario, demonstrating superior performance compared to baseline GNN models and traditional FEM methods. Our results show that MIXPINN reduces computational cost by an order of magnitude while maintaining high physical accuracy, making it a viable solution for real-time surgical simulation and robotic-assisted procedures.
\end{abstract}

\section{INTRODUCTION}
Spinal surgery is a complex and delicate procedure requiring precision and expertise. Simulation-based methods can improve surgical outcomes by enhancing preoperative planning \cite{dubrovin_virtual_2019}, surgical training \cite{elendu_impact_2024}, and real-time assistance in robotic procedures \cite{kawamura_real-time_2007}. Recent advancements in augmented reality (AR) \cite{avrumova_augmented_2022} and robotic-assisted interventions \cite{rehman_simulation-based_2013,lee_robotic-assisted_2024} have further emphasized the need for accurate, real-time tissue deformation modeling. However, achieving high accuracy while maintaining computational efficiency remains a fundamental challenge in physics-based medical simulations.

The Finite Element Method (FEM) is widely used in biomechanical simulations due to its ability to model tissue behavior with high accuracy \cite{naoum_finite_2021}. Despite its reliability, FEM suffers from high computational costs, hindering its real-time clinical adaptation. To address this limitation, learning-based approaches have emerged as a promising alternative. These data-driven models can approximate complex physical interactions accurately while significantly reducing computational requirements \cite{martinez-martinez_finite_2017}. Among data-driven models, Graph Neural Networks (GNNs) have recently demonstrated strong potential in predicting soft tissue deformations by leveraging mesh-based structures \cite{salehi_physgnn_2022,awwad_graph_2024}. However, existing methods primarily focus on soft tissue modeling and fail to accurately capture interactions between soft and rigid structures, which limits their applicability in scenarios requiring soft-rigid coupling, such as spinal surgery and ultrasound-guided interventions in orthopedics \cite{kowalska_ultrasound-guided_2014}.

\begin{figure}[t]
    \centering
    \includegraphics[width=\linewidth]{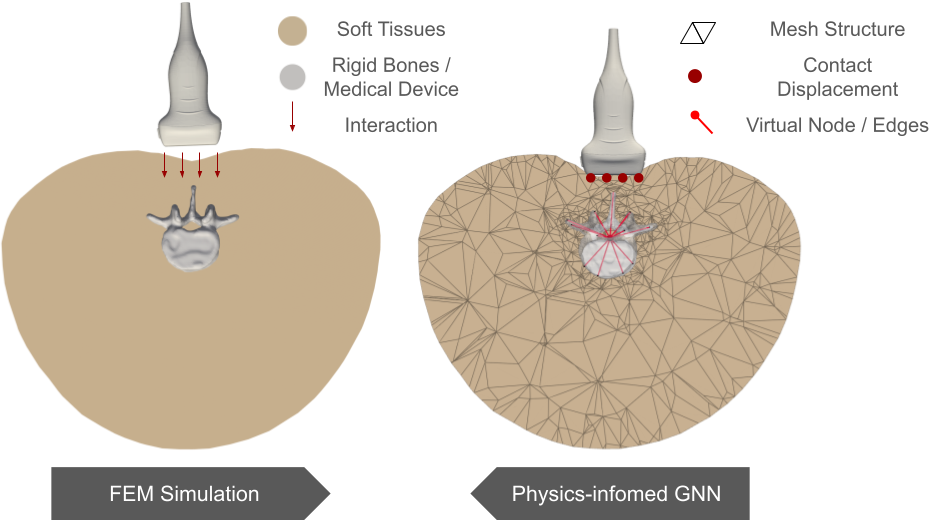}
    \caption{MIXPINN pipeline concept. For a FEM simulation scene of a rigid medical device interacting with multi-anatomy structure, we transform it to GNN training on a physics-informed mesh-based graph under the influence of surface contact displacements. The model aims to predict the internal deformations by considering the interplay between soft tissue and multiple rigid components.}
    \label{fig:teaser}
\end{figure}

To bridge this gap, we introduce MIXPINN, a physics-informed GNN framework designed for mixed-material simulations involving both soft tissue and rigid structures. We propose physics-informed graph augmentations, incorporating virtual nodes (VNs) and virtual edges (VEs) to improve rigid body constraints satisfaction. MIXPINN is trained on FEM-generated data and achieves real-time inference with sub-millimeter accuracy. As depicted in Fig. \ref{fig:teaser}, we evaluate our approach on an ultrasound (US) probe interaction scenario, where accurate deformation modeling is crucial.

Our contributions are summarized as follows:
\begin{itemize}
    \item First GNN-based approach for mixed-material simulation of soft-rigid interactions in a coherent structure.
	\item Physics-informed graph augmentation using virtual nodes and virtual edges to enforce rigid body constraints.
	\item Real-time inference capability, outperforming existing methods while preserving high-fidelity deformation accuracy ($<$ 0.3 mm error).
    \item First orthopedic simulation setup open-sourced in the Simulation Open Framework Architecture (SOFA) \cite{faure:hal-00681539}, providing a new benchmark for soft-rigid interaction modeling.
\end{itemize}

\section{Related Work}
\subsection{FEM-Based Medical simulation}
The Finite Element Method (FEM) is a standard approach for simulating biomechanical deformations by solving Partial Differential Equations (PDEs) based on material physics and Newtonian mechanics \cite{naoum_finite_2021}. FEM is widely used for surgical planning \cite{zhao_development_2024}, medical training and in surgical applications like soft tissue modeling \cite{freutel_finite_2014} and spinal biomechanics \cite{fasser_novel_2022, widmer_individualized_2020}. SOFA \cite{faure:hal-00681539} is a widely adopted open-source simulation framework that enables interactive multi-modal medical simulations. While FEM offers high accuracy, it comes with a significant computational cost, often taking minutes to hours depending on model complexity. This trade-off between accuracy and computational speed hinders its real-time surgical applications.

\subsection{Learning-Based Simulation for Tissue Deformation}
To address FEM’s computational limitations, data-driven methods have been developed to approximate tissue deformations using neural networks \cite{phellan_real-time_2021,martinez-martinez_finite_2017}. Recent studies explore deep learning models to accelerate soft tissue deformation predictions \cite{ying_wu_learning_2021}, but these methods primarily focus on homogeneous soft tissue deformation, lacking the ability to model rigid structures or mixed-material interactions.

\subsection{Graph Neural Networks for Mesh-Based Simulation}
Graph Neural Networks (GNNs) have shown promise in mesh-based physics simulations due to their ability to process non-Euclidean data structures \cite{asif_graph_2021}. \cite{pfaff_learning_2021} propose MESHGRAPHNETS and \cite{gladstone_mesh-based_2024} present two GNN surrogates for PDEs, both using encoders and decoders to transform mesh information to graph for GNN training. Although their methods are proven to be successful in multiple physics field, none of them incorporate rigid body dynamics, limiting their applicability in mixed-material simulations. 

\subsection{Physics-Informed Neural Network for Simulation}
Physics-Informed Neural Networks (PINNs) combine deep learning with physical priors to model complex PDEs \cite{raissi_physics-informed_2019} \cite{karniadakis_physics-informed_2021}. In the context of soft tissue simulation, PhysGNN \cite{salehi_physgnn_2022} encode physical properties like elasticity and external forces into neural networks. Similarly \cite{saleh_physics-encoded_2024} incorporate physics encoding into GNN to improve soft tissue deformations when contact with external rigid object. However, no existing PINN-based methods effectively capture soft-rigid interactions in a single graph-based framework.

\section{Method}
\subsection{Architecture}
MIXPINN is a data-driven model which incorporates Graph Attention Model (GAT) \cite{velickovic_graph_2018} to predict the displacement of all mesh nodes with different material properties under the influence of initial displacement of some surface mesh nodes. For a corresponding graph denoted $G = (V, E)$ where $V$ and $E$ represents the vertex set and edge set of the graph respectively, we use a matrix $\mathbf{X}:= [\mathbf{x}_1, \mathbf{x}_2, ..., \mathbf{x}_{|V|}]^\top \in \mathbb{R}^{d \times |V|}$ to represent the $d$-dimensional node features. For a node $i \in V$, its node feature embedding $\mathbf{x}_i$ is updated at each layer by aggregating information from its neighbors with itself weighted by attention, formulated as:
\begin{equation}
    \mathbf{x}'_i = \sum_{j \in N(i) \cup \{i\}} \alpha_{i,j} \, \Theta \mathbf{x}_j,
\end{equation}
where $j$ is a specific node in the neighborhood of $i$ given as $N(i)$. $\Theta$ is a learnable weight matrix of linear transform. For a graph with multi-dimensional edge features $\mathbf{e}_{i,j}$, the attention coefficients $\alpha_{i,j}$ between node $i$ and $j$ are computed as:
\begin{equation}
    \begin{split}
        \alpha_{i,j} &= \frac{\exp\left(\sigma^* \left( \mathbf{a}^\top \Theta \mathbf{x}_i + \mathbf{a}^\top \Theta \mathbf{x}_j + \mathbf{a}_e^\top \Theta_e \mathbf{e}_{i,j} \right)\right)}
        {\sum_{k \in N(i) \cup \{i\}} \exp\left( \sigma^* \left(\mathbf{a}^\top \Theta \mathbf{x}_i + \mathbf{a}^\top \Theta \mathbf{x}_k + \mathbf{a}_e^\top \Theta_e \mathbf{e}_{i,k} \right)\right)}
    \end{split}
\end{equation}
parametrized by weight vector $\mathbf{a}$ for node features and $\mathbf{a}_e$ for edge features. $\sigma^*:=$ LeakyReLU is the activation function used to add non-linearity into the computations. Multi-Head Attention (MHA) \cite{vaswani_attention_2023} enhances model capacity by learning multiple attention distributions in parallel. Each head calculates its own attention score $\alpha_{i,j}$ with different parameters. The attention output of each head is concatenated along the feature dimension and transformed by a weight matrix to give the final output. The final architecture of the MIXPINN model contains 8 two-headed GAT convolutional layers where each head has a hidden dimension of 256. 

\subsection{Input Features and Outputs}
Considering a use case which involves applying an US probe onto a 3D anatomical model of the human torso where this mesh-based model contains positions of the mesh nodes, each labeled with 0,1,2,... to identify the anatomy membership (See \ref{sec:sim} for details), Table \ref{tab:io} summarizes the input features and outputs of the GAT model in this use case. Both node features denoted $\mathbf{X}$ and edge features $\mathbf{E}$ are used as inputs feeding into the GAT model. The 14-dimensional node features are constructed from 4 components. First, $[\delta x_p, \delta y_p, \delta z_p]$ is the displacement of the mesh surface points which contact the US probe directly, and the rest of the points have a value of $[0,0,0]$. Then, the initial position of all mesh nodes centered around the origin in both Cartesian coordinates $[x, y ,z]$ and spherical coordinates $[r, \theta, \phi]$ are included. Finally, the rigid mask indicates the specific anatomy the node belongs to, in which 0 represents soft tissues and 1 to 5 represents vertebrae L1-L5 respectively. We include the one-hot encoding of the rigid mask $\mathbf{m}\in\{0, 1\}^5$ in the node features. Edge features of dimension 6 include the length of the edge $l$, which corresponds to the Euclidean distance between the two endpoints of the edge. Moreover, edge features also contain the one-hot encoding of its anatomy membership $\mathbf{m}_e\in\{0, 1\}^5$ similar to the node features. An edge is considered inside a specific rigid vertebra if both ends of the edge belong to that vertebra. Otherwise, it is considered as inside soft tissue with a label 0. The model output $\mathbf{Y}$ is the displacement of all mesh points in x, y and z directions, denoted $\delta x, \delta y, \delta z$.
\begin{table}[ht]
    \centering
    \caption{Input Features and Outputs of MIXPINN} \label{tab:io}
    \begin{tabular}{l|c}
        \hline
        Node Features ($\mathbf{X}$) & $\delta x_p, \delta y_p, \delta z_p, x, y, z, r, \theta, \phi, \mathbf{m}\in\{0, 1\}^5$ \\
        \hline
        Edge Features ($\mathbf{E}$) & $l, \mathbf{m_e}\in\{0, 1\}^5$ \\
        \hline
        Outputs ($\mathbf{Y}$) & $\delta x, \delta y, \delta z$ \\
        \hline
    \end{tabular}
\end{table}

\subsection{Physics-informed Graph Structure Augmentation}
\begin{figure}[ht]
    \centering
    \begin{minipage}{0.23\textwidth}
        \centering
        \includegraphics[width=\textwidth]{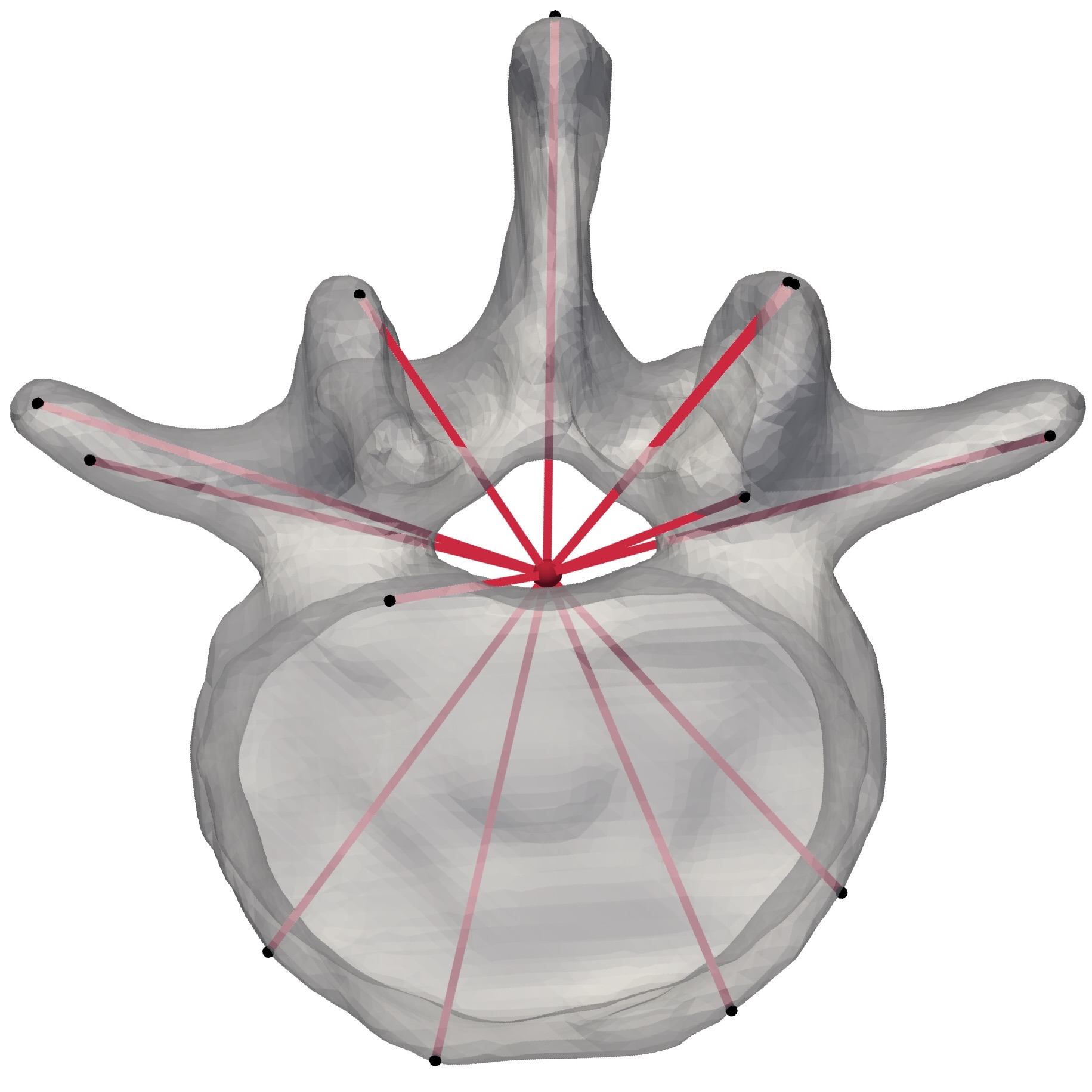}
    \end{minipage}
    \begin{minipage}{0.23\textwidth}
        \centering
        \includegraphics[width=\textwidth]{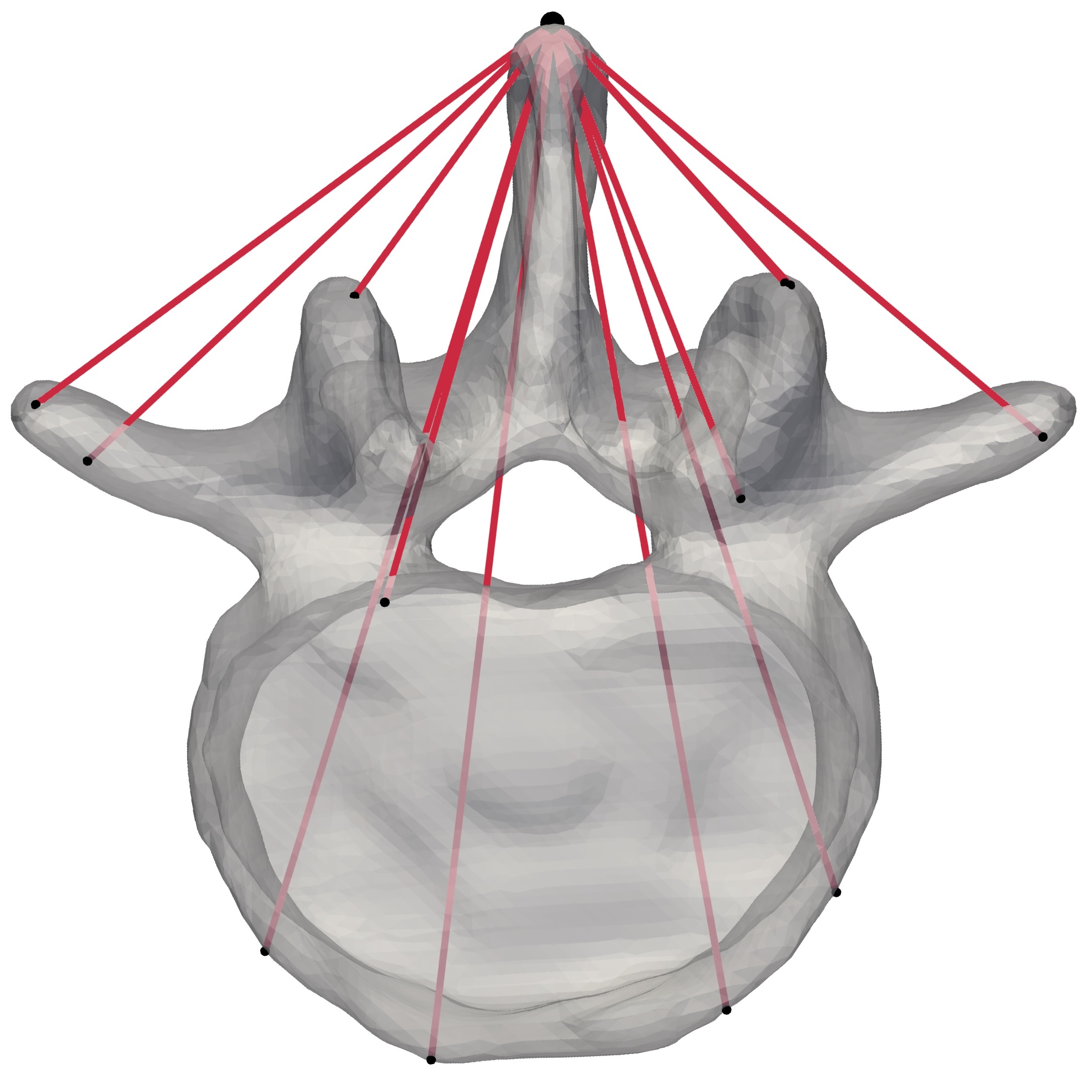}
    \end{minipage}
    \caption{Illustrations of the two different concepts VN (left) and VEs (right) graph structure augmentation. Black dots are points on the vertebra mesh. Red lines are edges added where on the left image red point is the virtual node added. Not all points and edges are shown.}
    \label{fig:vnve}
\end{figure}

A rigid object only undergoes translation and rotation as a whole and experiences no internal deformations. We propose two physics-informed approaches to augment the mesh-based graph structure in order to help the model comprehend this physical constraint. Fig. \ref{fig:vnve} shows the conceptual illustration of the virtual node-edges approach.

\paragraph{\textbf{Virtual Node} (VN) at the rigid body's center}
In FEM simulation, the mechanical state of a rigid object is modeled as a single point and all points in the object's mesh are mapped relatively to this representational position. Inspired from this, we append a virtual node for each rigid component in the mesh (in our case vertebrae L1-L5). It is positioned at the geometric center of its corresponding rigid part and is connected to all other nodes within the same rigid component. This design choice mimics the behavior of rigid objects in FEM simulations, where the center of mass are often chosen as the representative point. In this way, the virtual node acts as a central hub, integrating the messages from all the nodes in a specific rigid component. It also facilitates the communication between the rigid nodes by shortening the path from one rigid node to another. We construct node and edge features as well as output for the added virtual node as follows:
\begin{itemize}
    \item Initial Displacement: The initial displacement of the virtual nodes is set to zero, as they are positioned inside the body and do not contact the US probe directly.
    \item Rest Position: The position of each virtual node is computed as the mean position of all the nodes within its corresponding rigid component.
    \item Rigid Mask: The rigid mask for the virtual node has the same value as other points in the rigid part, effectively associating it with the correct vertebra.
    \item Edge Features: Similar to other edges in the graph, the edge features for the connections introduced between the virtual node and its neighboring rigid nodes include the edge length and the anatomy label of the virtual node.
    \item Output Displacement: The mean displacement of all the rigid nodes in the specific rigid part is used as the output of the virtual node.
\end{itemize}

\paragraph{\textbf{Virtual Edges} (VEs) connecting points in the rigid body}
While the virtual node approach requires additional node features and outputs for training, we propose a simplified alternative that only adds virtual edges to the graph. Instead of adding a new node, we designate an existing node within each vertebra as a representative reference point. All other nodes within the same rigid structure are then connected to this reference node through virtual edges, provided no edge already exists between them. The edge features of these newly added virtual edges follow the same construction principles as those in the VN approach. To ensure efficient information propagation, we select the reference node based on our simulation setup (Fig. \ref{fig:teaser}). Specifically, we choose the node closest to the surface deformation region where the external probe interacts with the body. This selection is crucial because this node is among the first to experience deformation, making it an ideal hub for message passing. By directly linking all nodes in the rigid structure to this reference node, MIXPINN facilitates faster and more effective gradient propagation throughout the network. This design reinforces rigid body constraints, ensuring that the vertebra moves as a cohesive unit. Ultimately, this enhances the model’s ability to predict deformations in complex biomechanical systems more accurately and efficiently as will be shown in results.

\subsection{Physics-informed Loss}
\paragraph{Mean Euclidean Error (MEE)}
Similar to PhysGNN \cite{faure:hal-00681539}, Mean Euclidean Error (MEE) is used as the main loss function for learning the trainable parameters, defined as:
\begin{equation}
    \mathbf{L}_\text{MEE} = \frac{1}{|V|} \sum_{i=1}^{|V|} \sqrt{(x_i - \hat{x}_i)^2 + (y_i - \hat{y}_i)^2 + (z_i - \hat{z}_i)^2},
    \label{eq:mee}
\end{equation}
where $|V|$ is the number of mesh nodes, $[x,y,z]$ is the ground truth displacements approximated by FEM and $[\hat{x}, \hat{y}, \hat{z}]$ is the predicted displacements from MIXPINN.

\paragraph{\textbf{Rigid Edge Loss} (REL)}
A set of N points forms a rigid body if the distance between any 2 points is fixed. This can be expressed mathematically as:
\begin{equation}
    d_{ij} := \|\mathbf{r} _i- \mathbf{r} _j\|_2= c_{ij} = \text{constant.}  
\end{equation}
$\mathbf{r}_i$ and $\mathbf{r}_j$ are positions of any two points on the rigid body (including virtual nodes). $d_{ij}$ is the Euclidean distance between them. To enforce the rigid body constraint, MIXPINN introduces the rigid edge loss (REL) that penalizes any deviation from the constant edge length $c_{ij}$ between nodes within the same rigid part. REL is formulated as follows by adopting the Mean Squared Error (MSE) loss function:
\begin{equation}
    \mathbf{L}_\text{REL}  = \frac{1}{|E_\text{rigid}|} \sum_{i=1}^{|E_\text{rigid}|} (c_i - d_i)^2 = \frac{1}{|E_\text{rigid}|} \sum_{i=1}^{|E_\text{rigid}|} (l_i - \hat{l}_i )^2,
\end{equation}
in which $|E_\text{rigid}|$ is the number of edges in vertebrae. $l$ is the constant edge length obtained from edge features (Table \ref{tab:io}) and $\hat{l}$ is the computed edge length with the predicted deformation. Minimizing REL incentivize the model to learn displacements that preserve the integrity of the rigid structure, even when the surrounding soft tissues deform.

\section{Experiment}
\subsection{Data Acquisition}
\label{sec:sim}
\begin{figure*}
    \centering
    \includegraphics[width=\textwidth]{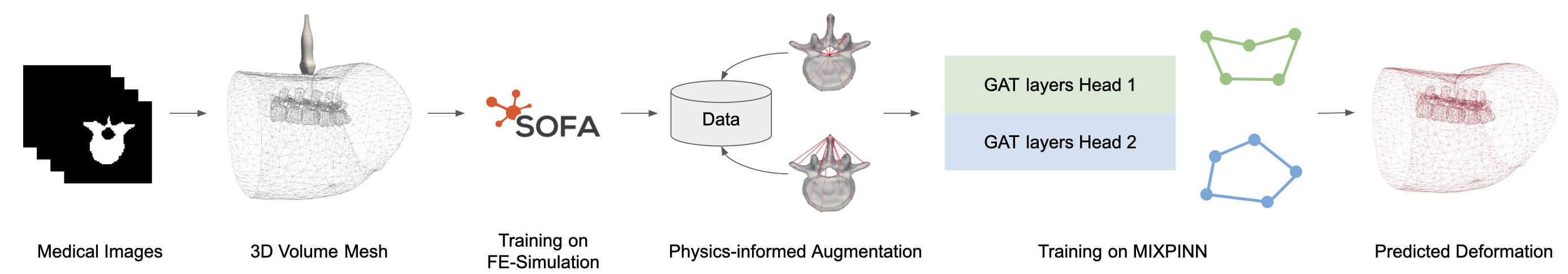}
    \caption{Experiment pipeline of MIXPINN. Generate 3D mesh from medical images and train on FE-Simulation with it to produce the data for GNN training. Then perform physics-informed graph structure augmentation on inputs and feed it into two-headed GAT models to predict the mesh deformation.}
    \label{fig:mixpinn}
\end{figure*}

The data for training the GAT model are generated from running FEM experiments using SOFA \cite{faure:hal-00681539}. In the simulation, an Ultrasound (US) probe compress the body back at different locations, rotation angles and depth. Fig. \ref{fig:mixpinn} provides an overview of the experiment pipeline for multi-anatomy simulations with MIXPINN model.

\paragraph{Mesh Generation from Medical Images}
Six segmentations from a human body Computed Tomography (CT) image are obtained from TotalSegmentator dataset~\cite{wasserthal_totalsegmentator_2023}, namely, the main body trunk and lumbar vertebra L1 to L5. These images have $311 \times 311 \times 431$ 3D resolution with 1.5 mm voxel size. Using marching cubes, they are processed and combined to produce a multi-tissue triangular surface mesh for the body lumbar region. Then the surface mesh is used to generate the 3D tetrahedral mesh using The Computational Geometry Algorithms Library (CGAL) plugin in the SOFA framework with parameters cellRatio=2, cellSize=40, facetSize=30, facetAngle=30 and facetApproximation=1. Each mesh point and edge is labeled with the anatomy segment where 0 indicates soft tissue and 1 to 5 corresponds to vertebra L1-L5 respectively. Table \ref{tab:mesh} provides detailed statistics of the mesh generated.
\begin{table}[ht]
    \centering
    \caption{Statistics of mesh points and edges in each anatomy label} \label{tab:mesh}
    \begin{tabular}{c|c|c}
        \hline
        Anatomy (label) & Mesh Points & Mesh Edges \\
        \hline
        Soft Tissue (0) & 5533 & 74896 \\
        Vertebra L1 (1) & 516 & 3372 \\
        Vertebra L2 (2) & 676 & 4634 \\
        Vertebra L3 (3) & 655 & 4350 \\
        Vertebra L4 (4) & 702 & 4766 \\
        Vertebra L5 (5) & 664 & 4680 \\
        \hline
        Total & 8746 & 96698\\
        \hline
    \end{tabular}
\end{table}

\paragraph{Finite Element (FE)-Modeling and Simulation}
The mechanical properties of soft tissues are characterized by the hyper-elastic Stable Neo-Hookean flesh model \cite{smith_stable_2018}, which is parametrized by a Poisson's ratio of 0.45 and Young's modulus of 25400 (Pa) \cite{miyamori_differences_2023}. Based on the significant stiffness discrepancy between soft tissues and spine, the lumbar vertebrae are modeled as rigid bodies \cite{vena_finite_2005}. This is achieved by subsetting the main body mesh and rigidifying the mesh nodes in each rigid vertebra separately. In this way, a multi-anatomy model is established. This mixed-material FE-model contains a deformable part and a rigid part which further divides to 5 sub-components. The US probe is also a rigid object in the simulation constructed from a Telemed L12 linear ultrasound transducer mesh \cite{lasso_plus_2014}. To set up the boundary condition, some mesh nodes in the body anterior surface are fixed to support and stabilize the simulation. The initial probe positions are sampled from a structured mesh grid on the body back with 10 mm interval which produces 132 distinct positions. At each location, the US probe is positioned with four orientations corresponding to 0°, 45°, 90°, and 135° rotations. With each sampled pose, the probe moves at a uniform speed of 10 mm/s towards body back. We use the collision based contact response in SOFA with the friction coefficient set to 1 for a non-slippery contact. The simulation is ran for 20 time steps which corresponds to the probe compressed 2 cm into the body. For each time step of 0.1 s, the quasi-static state of the system is resolved and computed. The displacements of all mesh points are exported. In this way, a total of 9796 experiment samples are created for training and validating the model.

\subsection{Held-out Experiment}
To evaluate the model’s ability to generalize deformation predictions to unseen US probe position and pose, a specific testing strategy was used. The 132 simulation experiments of unique probe position are split by a train, validation, test ratio of $[0.7, 0.2, 0.1]$. Their respective probe poses (from 4 rotations) are also split accordingly. Table \ref{tab:data} reports the detailed statistics of the dataset created for GNN training. During training, a batch size of 4 is used and the data will be randomly shuffled after each epoch.
\begin{table}[ht]
    \centering
    \caption{Statistics of the dataset used for GNN training generated from FE-simulation} \label{tab:data}
    \begin{tabular}{c|c|c|c|c}
        \hline
         & Train & Validation & Test & Total\\
        \hline
        Probe Position & 93 & 26 & 13 & 132 \\
        Probe Pose & 372 & 104 & 52 & 528 \\
        Generated Sample & 6876 & 1963 & 957 & 9796 \\
        \hline
    \end{tabular}
\end{table}

\subsection{Evaluation Metrics}
To evaluate the performance of the model quantitatively, Mean Euclidean Error (MEE) as in Eq. \ref{eq:mee}, Mean Absolute Error (MAE) and Mean Squared Error (MSE) which evaluate the discrepancy between predicted displacements with ground-truth displacements from the FE-simulation are measured. Their equations are provided below:
\begin{equation}
\text{MAE} = \frac{1}{|V|} \sum_{i=1}^{|V|} \left( (x_i - \hat{x}_i) + (y_i - \hat{y}_i) + (z_i - \hat{z}_i) \right) \label{eq:mae}
\end{equation}
\begin{equation}
    \text{MSE} = \frac{1}{|V|} \sum_{i=1}^{|V|} \left( (x_i - \hat{x}_i)^2 + (y_i - \hat{y}_i)^2 + (z_i - \hat{z}_i)^2 \right) \label{eq:mse}
\end{equation}

To investigate the prediction accuracy for the soft tissues and rigid vertebrae separately, their respective MEEs, averaged over the number of points on rigid vertebrae and the number of points on soft tissues, are also computed. In addition, Rigid Edge Error (REE), similar to REL but measures the absolute change in rigid edge length is reported to assess the integrity of the rigid vertebrae.

\subsection{Baselines}
The predictions made by the proposed model is compared with PhysGNN \cite{salehi_physgnn_2022} against the approximations by FEM. The performance of a simple GAT model with 8 single-head attention layers is also used as a baseline.

\subsection{Optimization and Regularization}
To minimize the loss, the trainable parameters are optimized by AdamW optimizer with a weight decay of 0.01. Starting with a learning rate of 0.0005, the learning rate is decayed by a factor of 0.1 to a minimum value of 1e-8 when the validation loss does not improved in 5 epochs. Early stopping are deployed to prevent over-fitting, the training process stops if the validation loss did not decrease after 15 epochs.

\subsection{Runtime Profiling}
In order to compare the efficiency of MIXPINN with FEM simulation in generating displacement predictions, their runtime are measured. For FEM simulation, we use the built-in SOFA timer to extract the records from 20 animation steps in 132 simulation experiments and average the time. The inference time of best performing MIXPINN model are recorded at test stage and averaged over the number of test samples. The FEM simulations are carried out on AMD EPYC 64-Core Processor CPU and various GNN models are trained on the same CPU with one NVIDIA GeForce RTX 4090 GPU.

\section{Results and Discussion}
\subsection{Quantitative Evaluation}
Table \ref{tab:results} reports the performance of MIXPINN and baseline models. We can see that MIXPINN significantly outperforms PhysGNN in all metrics, especially in Rigid MEE and REE. This proves the efficacy of the MIXPINN model in predicting soft-rigid mixed-material deformations while complying the physics of rigid components.
\begin{table*}[t]
\caption{Comparison of deformation prediction accuracy between MIXPINN and baselines. Best results are in \textbf{bold}.}\label{tab:results}
\begin{center}
\begin{tabular}{c|c|c|c|c|c|c}
\hline
Model & MEE $[\mathrm{mm}]$ $\downarrow$ & MAE $[\mathrm{mm}]$ $\downarrow$ & MSE $[\mathrm{mm}^2]$ $\downarrow$ & Rigid MEE $[\mathrm{mm}]$ $\downarrow$ & Soft MEE $[\mathrm{mm}]$ $\downarrow$ & REE $[\mathrm{mm}]$ $\downarrow$ \\
\hline
PhysGNN & 0.7358  & 0.3593 & 0.3604 & 0.8209 & 0.6864 & 0.1238 \\
Simple GAT & 0.3847 & 0.1905 & 0.1266 & 0.3874 & 0.3831 & 0.0723 \\
MIXPINN & \textbf{0.2588} & \textbf{0.1304} & \textbf{0.0670} & \textbf{0.1663} & \textbf{0.3126} & \textbf{0.0326} \\
\hline
\end{tabular}
\end{center}
\end{table*}

\subsection{Qualitative Evaluation}
\begin{figure}[ht]
    \centering
    \includegraphics[width=\linewidth]{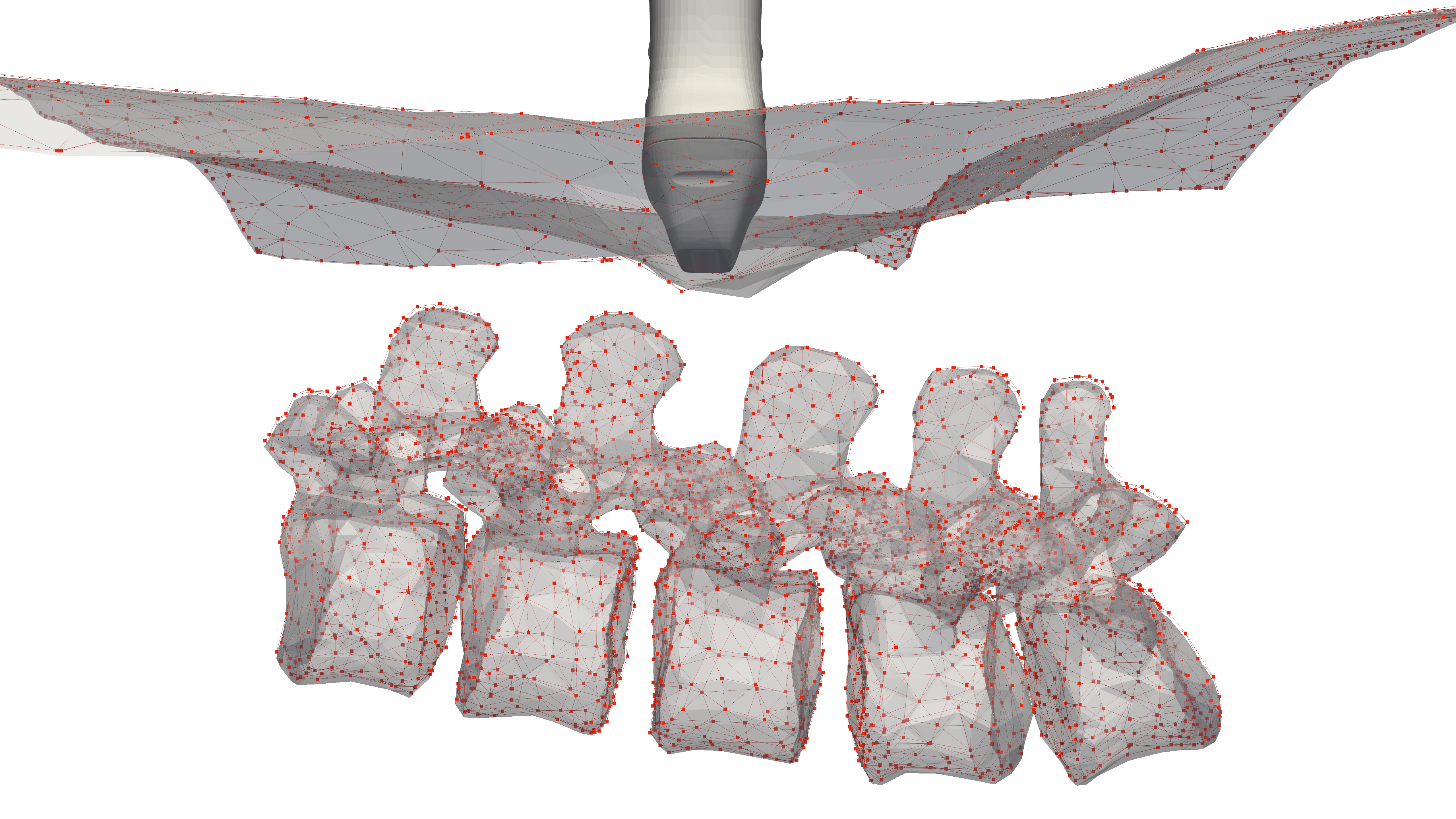}
    \caption{Visualization of MIXPINN's predictions (red wire-frame) with ground-truth from FE-simulation (gray mesh). For clarity, only a section of the meshes are shown.}
    \label{fig:pred}
\end{figure}

Fig. \ref{fig:pred} shows a deformed mesh predicted by MIXPINN together with the ground-truth mesh computed in the simulation (grey). We can observe that the red wire-frame of the predicted mesh align with the contour of the true mesh almost perfectly. This is consistent with the numerical results of less than 1 mm errors. 

\subsection{Ablation Study}
To identify the best-performing MIXPINN model, we evaluate various MIXPINN configurations that incorporate different combinations of GAT layers and physics-informed components. 
\begin{table*}[t]
    \centering
    \caption{Ablation study: performance of MIXPINN with different configurations. Best results are in \textbf{bold}.} \label{tab:ablation}
    \setlength{\tabcolsep}{4pt}
    \begin{tabular}{c|ccccc|cccccc}
        \hline
        Experiment & \multicolumn{5}{c|}{Configuration} & \multicolumn{6}{c}{Metric} \\
        & Head & Edge Feature & REL & VN & VE & MEE $[\mathrm{mm}]$ & MAE $[\mathrm{mm}]$ & MSE $[\mathrm{mm}^2]$ & Rigid MEE $[\mathrm{mm}]$ & Soft MEE $[\mathrm{mm}]$ & REE $[\mathrm{mm}]$ \\
        \hline
        1 & 1 & & & & & 0.3847 & 0.1905 & 0.1266 & 0.3874 & 0.3831 & 0.0723 \\
        2 & 2 & & & & & 0.3395 & 0.1686 & 0.1003 & 0.3456 & 0.3359 & 0.0750 \\
        \hline
        3 & 2 & $\checkmark$ & & & & 0.3335 & 0.1657 & 0.0990 & 0.3356 & 0.3323 & 0.0736 \\
        4 & 2 & $\checkmark$ & $\checkmark$ & & & 0.3259 & 0.1616 & 0.0961 & 0.3418 & 0.3166 & 0.0636 \\
        5 & 2 & $\checkmark$ & & $\checkmark$ & & 0.2735 & 0.1376 & 0.0710 & 0.1775 & 0.3293 & 0.0323 \\
        6 & 2 & $\checkmark$ & & & $\checkmark$ & 0.2721 & 0.1368 & 0.0715 & 0.1732 & 0.3296 & 0.0362 \\
        7 & 2 & $\checkmark$ & $\checkmark$ & $\checkmark$ & & 0.2759 & 0.1386 & 0.0724 & 0.1794 & 0.3320 & \textbf{0.0256} \\
        8 & 2 & $\checkmark$ & $\checkmark$ & & $\checkmark$ & 0.2750 & 0.1383 & 0.0721 & 0.1779 & 0.3314 & 0.0270 \\
        \hline
        9 & 2 & & $\checkmark$ & & & 0.3394 & 0.1681 & 0.0992 & 0.3537 & 0.3312 & 0.0641 \\
        10 & 2 & & & $\checkmark$ & & \textbf{0.2588} & \textbf{0.1304} & \textbf{0.0670} & \textbf{0.1663} & \textbf{0.3126} & 0.0326 \\
        11 & 2 & & & & $\checkmark$ & 0.2800 & 0.1408 & 0.0733 & 0.1885 & 0.3331 & 0.0474 \\
        12 & 2 & & $\checkmark$ & $\checkmark$ & & 0.2674 & 0.1344 & 0.0699 & 0.1701 & 0.3239 & 0.0258 \\
        13 & 2 & & $\checkmark$ & & $\checkmark$ & 0.2830 & 0.1424 & 0.0737 & 0.1888 & 0.3377 & 0.0348 \\
        \hline
    \end{tabular}
\end{table*}
Table \ref{tab:ablation} suggests that when edge features are explicitly passed into GAT convolutional layers (Experiment 3-8), the model with VE (Experiment 6) achieves a best MEE score of 0.2721 mm. On the other hand, if the GAT layers take only node features as input, and edge features are used solely by REL (Experiment 9-13), the addition of VN itself together with MHA are sufficient to give a MEE of 0.2588 mm. In fact, this is the configuration with the best performance for our MIXPINN model, which again proves the efficacy of VN augmentation. The ablation studies also help us to better understand the intricacies of our model and the influence of distinct configurations. By selectively activating the physics-informed loss and physics-informed graph structure augmentations, individually and in combination, we could assess their individual and combined effects on prediction accuracy. While the addition of REL improves the model's compliance to rigid body physical constraint measured by REE, it does not always help with the overall performance. One possible explanation is that, the displacement and rotation caused by external interaction with US probe are limited in the experiment, so that REE are naturally small. Thus, with the help of physics-informed structure augmentation, it is sufficient for GAT to learn the rigid behaviors implicitly.

\subsection{Runtime}
The runtime evaluation of the FEM simulation and MIXPINN are reported in Table \ref{tab:time}. While FEM simulation need around 1 s, MIXPINN uses 10-folded less time to predict the displacement of all points in the mesh. We also notice in the simulation that the time required to compute outputs increases with the mesh structure's complexity and the number of constraints created by external interactions with the US probe. In contrast, with a pre-trained model, MIXPINN uses relatively constant time, invariant to the extent of deformation. This suggests that our model is suitable for real-time applications in surgical systems where efficiency and accuracy are both necessary. 
\begin{table}[ht]
    \centering
    \caption{Average runtime of one mesh deformation prediction} \label{tab:time}
    \begin{tabular}{c|c}
        \hline
        Method & Time [ms] \\
        \hline
        FEM Simulation & 1009.63 \\
        MIXPINN & 73.24 \\
        \hline
    \end{tabular}
\end{table}

\subsection{Discussion}
By providing real-time yet accurate deformation prediction on multi-anatomy tissues, our method can be deployed to enhance intra-operative surgical planning or used for tool trajectory predictions. We plan to improve our method in a few possible areas. For example, the model can be modified to incorporate time series data to simulate beyond quasi-static scenario. On the other hand, the model performance can be further validated by conducting physical experiments on phantom or cadaver. The method can also be benchmarked on a wider range of clinical scenarios.

\section{CONCLUSIONS}
In this work, we propose MIXPINN, a novel physics-informed GNN framework for simulating soft-rigid interactions in mixed-material biomechanical systems. Unlike existing learning-based models that primarily focus on soft tissue deformations, MIXPINN integrates Virtual Nodes (VNs) and Virtual Edges (VEs) to enforce rigid body constraints, ensuring physically plausible deformations in a computationally efficient manner. Our results demonstrate that MIXPINN achieves state-of-the-art performance with a 10 times speedup over FEM while maintaining sub-millimeter accuracy. Through extensive evaluations on ultrasound probe interaction simulations, we showcase its potential in surgical training, robotic-assisted procedures, and real-time simulation environments. Our work establish a strong benchmark for efficient, high-fidelity biomechanical simulation, bridging the gap between data-driven models and physics-based constraints for real-world medical applications.

\section*{ACKNOWLEDGMENTS}
This work is part of the "Learn to learn safely" project funded
by a grant of the Hasler foundation (grant nr: 21039). Furthermore, we extend our sincere thanks to Hugo Talbot for his valuable consulting on the SOFA Framework.

\bibliographystyle{IEEEtran}
\bibliography{mybibfile}

\end{document}